%% file: root.tex
\documentclass[letterpaper, 10 pt, conference]{ieeeconf}  %

\IEEEoverridecommandlockouts                              %

\title{\LARGE \bf
TOLiD: Bridging the Architecture Gap in Vision Foundation Model to LiDAR Pretraining via Token Lifting for Distillation }

\author{Sutharsan Mahendran$^{1,2}$, Darshana Priyasad$^{1}$, Kaushik Roy$^{2}$, Tharindu Fernando$^{1}$, \\ Sridha Sridharan$^{1}$,  Clinton Fookes$^{1}$, Peyman Moghadam$^{1,2}$ \thanks{$^{1}$SAIVT Group, School of Electrical Engineering and Robotics, Queensland University of Technology, Australia.  E-mails: {\tt\footnotesize \{s2.mahendren, s.sridharan, c.fookes, peyman.moghadam\}@qut.edu.au}} \thanks{$^{2}$CSIRO~Robotics, CSIRO, Australia. E-mails: {\tt\footnotesize \{sutharsan.mahendran, kaushik.roy, peyman.moghadam\}@csiro.au}}}

\usepackage[T1]{fontenc}
\usepackage[utf8]{inputenc}
\usepackage{amssymb}
\usepackage{amsmath}
\usepackage{graphicx} 
\usepackage{xcolor}
\usepackage{cite}
\usepackage{booktabs}
\usepackage{multirow} 
\usepackage{array}
\usepackage{colortbl}
\usepackage{lipsum}
\usepackage{subcaption}
\usepackage{adjustbox}
\newcommand{\coolname}{\textit{TOLiD}}
\usepackage{comment}

\usepackage[bookmarks=true]{hyperref}

\definecolor{rowgray}{gray}{0.9}
\definecolor{roworange}{RGB}{255, 235, 215}
\definecolor{dinos28k}{RGB}{220,252,231}   
\definecolor{dinos55k}{RGB}{187,247,208}  
\definecolor{dinol28k}{RGB}{253,235,218}  
\definecolor{dinol55k}{RGB}{248,215,190}   
\definecolor{dinol79k}{RGB}{255, 235, 215} 
\definecolor{rulegray}{RGB}{180,180,180}

\newcommand{\eg}{\emph{e.g.},}
\newcommand{\ie}{\emph{i.e.},}

\usepackage{array}
\newcolumntype{P}[1]{>{\raggedright\arraybackslash}p{#1}}
\newcolumntype{C}[1]{>{\centering\arraybackslash}p{#1}}

\begin{document}

\bstctlcite{IEEEexample:BSTcontrol}

\maketitle
\thispagestyle{empty}
\pagestyle{empty}

\everypar{\looseness=-1}

\input{Section/00-abstract}
\input{Section/01-introduction}
\input{Section/relatedwork}

\input{Section/method}

\input{Section/experiments}

\input{Section/results}

\input{Section/conclusion}

{\small
\bibliographystyle{IEEEtran}
\bibliography{IEEEabrv,IEEEexample}
}
\end{document}

%% file: Section/00-abstract.tex
\begin{abstract}
Cross-modal distillation from Vision Foundation Models (VFMs) to LiDAR backbones has recently emerged as a self-supervised pretraining strategy that reduces reliance on dense point-wise annotation for 3D scene understanding. 
However, existing distillation pipelines typically treat the VFM as a frozen feature source and train a heterogeneous 3D backbone to match fixed image embeddings, forcing the student to bridge both the modality gap and the cross-architecture gap between dense ViT token representations and sparse 3D encoders. We propose \coolname{}, a self-supervised pretraining method for LiDAR representation learning that addresses this gap by coupling a LiDAR backbone with a student Vision Transformer (ViT) initialized from a frozen VFM teacher and applying supervision over compatible patch-token representations. \coolname{} converts the set of point features within each image patch frustum into a token using \emph{Frustum Pooling} followed by \emph{Frustum Attention}, and performs token-level distillation with visibility masking. For LiDAR-only deployment, we lift token features back to per-point representations using masked bilinear sampling to avoid patches that have limited LiDAR points. We extensively evaluate \coolname{} on five heterogeneous LiDAR datasets and four cross-sensor adaptation pairs, demonstrating improved transfer with frozen backbones and lightweight heads.

\end{abstract}

%% file: Section/01-introduction.tex
\section{Introduction}
\label{sec:intro}

3D semantic segmentation is a core capability for autonomous robots and self-driving cars, enabling 3D scene understanding for navigation, mapping, and safe interaction~\cite{behley2019semantickitti,vidanapathirana2025wildscenes, ancha2024deep, chen2019suma++, sun2020pointmoseg}. In many robotics systems, 3D scene understanding relies primarily on LiDAR, as it provides metric geometric measurements that are largely invariant to illumination and support reliable long-range perception. 
 
LiDAR-based learning models must operate across heterogeneous sensor suites that differ in beam count, field of view, point density, and scanning pattern, and these variations induce cross-sensor domain shifts when models are transferred between platforms. Addressing such sensor gaps by collecting new annotations is often impractical, since dense point-wise labelling of LiDAR scans is labour-intensive and costly~\cite{behley2019semantickitti}. 

While 2D perception tasks have benefited substantially from Vision Foundation Models (VFMs) pre-trained on large-scale datasets~\cite{shang2024theia, hausler2025pair}, progress for LiDAR-based tasks has been comparatively constrained by the absence of suitable 3D foundation models and by the modality gap that arises when utilizing VFMs for LiDAR representations~\cite{pmlr-v305-jung25c, hindel2025label}.
To address these limitations, recent work has leveraged VFMs as teachers and distilled their representations directly to LiDAR backbones as a self-supervised pretraining step, followed by a finetuning stage that requires substantially less labeled 3D data. These VFM-to-LiDAR cross-modal distillation methods typically follow a two-stage pipeline. In the pretraining stage, synchronized camera–LiDAR pairs are used to establish 2D–3D correspondences: LiDAR points~\cite{liu2021learning, puy2024three}~(or superpoints~\cite{sautier2022image, liu2023segment}) are projected into the image plane and associated with dense VFM features (patch tokens or intermediate feature embeddings). Then LiDAR encoder is trained to predict or align to these VFM embeddings, often through feature regression~\cite{puy2024three, govindarajan2025cleverdistiller}, or contrastive objectives~\cite{10802675, liu2021learning, sautier2022image} that encourage the LiDAR encoder to mimic the teacher’s embedding structure. 

Despite their effectiveness, current VFM-to-LiDAR cross-modal distillation pipelines face a fundamental limitation. They often distill from very large ViT teachers into comparatively small 3D backbones with substantially different architectures and feature parameterizations, forcing the student to bridge both the cross-modal gap and the cross-architecture gap simultaneously. This mismatch can lead to information loss or unstable transfer, where only coarse cues are retained, and fine-grained semantics that matter for the point-wise semantic segmentation task do not distill reliably, particularly under strong sensor or environmental shifts.

To tackle these limitations, we propose \coolname{}, a self-supervised pretraining method for LiDAR representation learning that addresses the architectural gaps in existing cross-modal distillation pipelines. Instead of distilling VFM features directly into a heterogeneous 3D LiDAR backbone, \coolname{} couples a LiDAR backbone with a student Vision Transformer (ViT) initialized with a frozen VFM teacher, so that supervision is applied over compatible patch-token representations rather than across heterogeneous architectures. Distillation is formulated on the teacher's transformer outputs, using patch tokens (and a global [CLS] token) as self-supervision targets. 

To make LiDAR features compatible with patch tokens, \coolname{} represents each image patch by its corresponding 3D frustum and converts the set of point features in that frustum into a single token. We propose two complementary mechanisms for this conversion: \emph{Frustum Pooling} (FP) provides a simple and stable aggregation that maps variable-size point sets into tokens, while \emph{Frustum Attention} (FA) allows each token to selectively emphasise the most informative points within its frustum via query-based weighting, which is important when point density, occlusion, and sampling patterns vary across varieties of LiDAR sensors. Finally, \coolname{} supports LiDAR-only deployment by lifting token features back to per-point representations. To avoid introducing spurious signals from image patches with no LiDAR support, we use masked bilinear sampling that restricts interpolation to supported tokens and renormalises the weights.  

We pretrain \coolname{} on five heterogeneous camera-LiDAR datasets with different sensor configurations. 
\coolname{} exhibits stronger scaling with pretraining diversity. 
When evaluated with a frozen backbone and a lightweight segmentation head, the average performance gap over the prior state-of-the-art widens from {+2.0\%} to {+4.2\%} mIoU as pretraining data increases more than double from 169k to 342k frames. 
The self-supervised representations learned by \coolname{} transfer directly under domain shift. A lightweight MLP head trained on a labelled source dataset achieves SOTA accuracy on  target domain across four cross-sensor pairs. 
These results demonstrate that coupling the VFM student with the LiDAR backbone yields deployment ready representations.

%% file: Section/relatedwork.tex
\begin{figure*}[!t]
    \centering
    \includegraphics[width=0.75\linewidth]{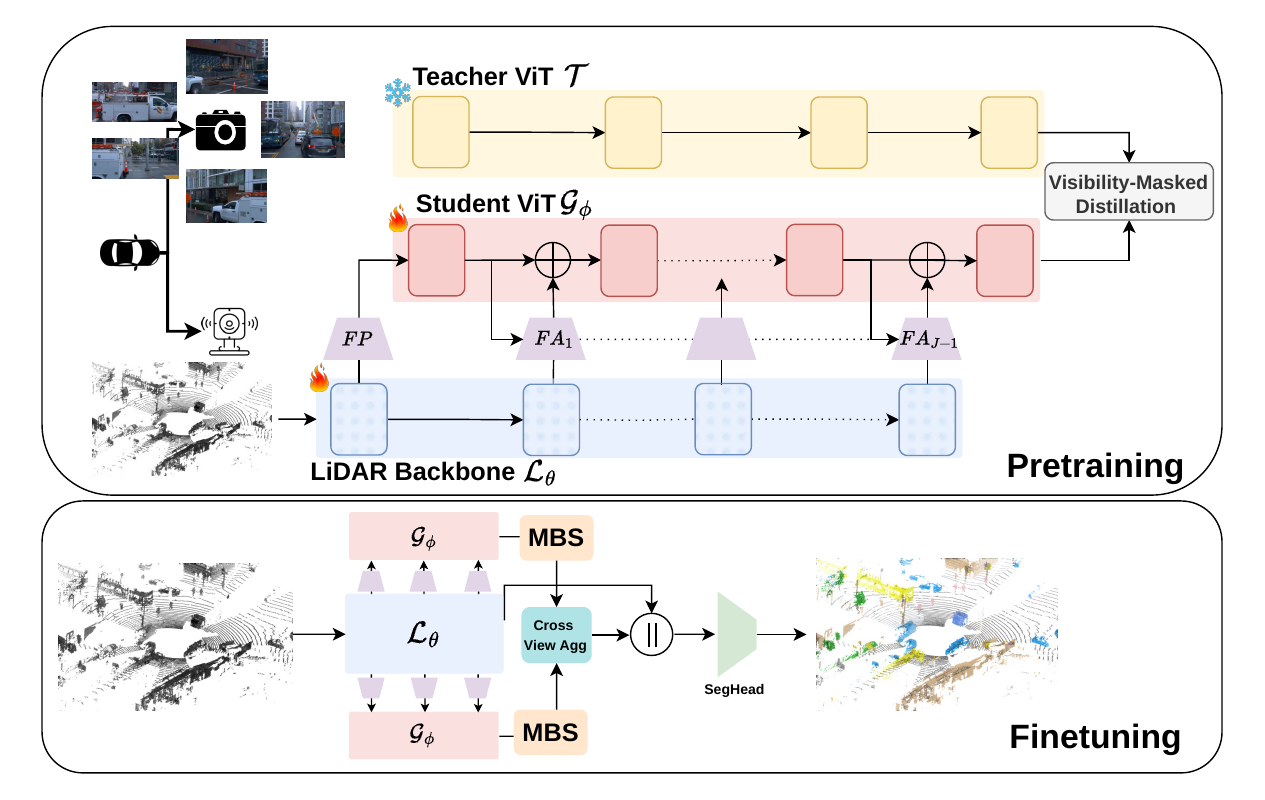}  
    \caption{\textbf{Overview of the \coolname{}.} During pretraining (\textit{top}), $\mathcal{L}_\theta$ converts the point features into patch tokens via \emph{Frustum Pooling (FP)} and \emph{Frustum Attention (FA)} at $J$ stages and integrates into the $\mathcal{G}_\phi$; visibility-masked distillation aligns them with the frozen teacher $\mathcal{T}$. During finetuning (\textit{bottom}), token features are lifted back to per-point representations via Masked Bilinear Sampling (MBS) across all views and concatenated with $\mathcal{L}_\theta$ features for segmentation.}
    \label{fig:complete_model}
\end{figure*}

\section{Related Work}
\label{sec:related_works}
\subsection{Self-Supervised LiDAR Representation Learning}
Self-supervised pretraining methods learn transferable 3D representations from unlabeled point clouds using contrastive objectives~\cite{pointpng} and masked reconstruction~\cite{11127773}.
While these approaches provide effective initializations, they are inherently limited to geometric and structural cues and lack the rich semantic abstractions encoded in large-scale pretrained image models, motivating cross-modal distillation. 

\subsection{VFM-to-LiDAR Cross-Modal Distillation}
Distilling pretrained 2D vision backbones into 3D LiDAR networks has emerged as an effective self-supervised pretraining strategy that leverages synchronized camera-LiDAR pairs without point-wise annotations. PPKT~\cite{liu2021learning} introduced pixel-to-point contrastive alignment but is primarily suited to dense indoor RGB-D settings. To handle the sparsity and occlusion of outdoor LiDAR, SLidR~\cite{sautier2022image} proposed superpixel-level contrast, though class-agnostic superpixels introduce semantic conflicts. Seal~\cite{liu2023segment} addresses this by leveraging foundation-model semantic regions, while CLIP2Scene~\cite{chen2023clip2scene} exploits CLIP's text-aligned embeddings to guide sampling. 

More recently, ScaLR~\cite{puy2024three} and LargeAD~\cite{kong2025largead} demonstrated that pretraining across multiple heterogeneous LiDAR datasets with direct feature-level alignment can improve cross-sensor generalization. However, these methods still treat the VFM as a frozen feature source and train a heterogeneous 3D backbone to match fixed image embeddings, forcing the student to bridge both the modality gap and the cross-architecture gap between dense ViT token representations and sparse 3D encoders. This restricts distillation to final feature alignment and limits how effectively the structured representations encoded in modern VFMs transfer as pretraining data grows in scale and sensor diversity. 
More broadly, prior work in the image domain has shown that transferring knowledge across such heterogeneous architectures is inherently difficult~\cite{raghu2021dovit, hao2023one}, as differences in receptive field, block structure, and normalization yield incompatible feature spaces~\cite{zhao2023cumulative}.
In contrast, \coolname{} addresses this gap by coupling a LiDAR backbone with a student ViT initialized from a frozen VFM teacher, so that supervision is applied over compatible patch-token representations rather than across heterogeneous architectures.

%% file: Section/method.tex
\section{Methodology}
\label{sec:method}
Existing VFM-to-LiDAR cross-modal distillation methods treat the VFM as a frozen feature source and train a heterogeneous 3D backbone to match fixed image embeddings~\cite{puy2024three, govindarajan2025cleverdistiller, sautier2022image}, requiring the LiDAR network to bridge both the modality gap and the cross-architecture gap. In contrast, \coolname{} couples a LiDAR backbone $\mathcal{L}_\theta$ with a student Vision Transformer (ViT) $\mathcal{G}_\phi$ initialized with a frozen VFM teacher, $\mathcal{T}$ (Fig.~\ref{fig:complete_model}). 
The LiDAR backbone provides point-wise features that are projected into calibrated camera views and converted into ViT-compatible patch tokens, a process that requires: (i) projecting 3D point features into each camera view using the correspondences, (ii) aggregation of variable-size frustum sets into fixed-dimensional tokens, and (iii) multi-stage alignment across multiple stages within ViT blocks. We implement this through \emph{Frustum Pooling} at the ViT input and \emph{Frustum Attention} at subsequent stages. This enables distillation to be formulated directly over the teacher's transformer representations.

\subsection{Preliminaries and Notations}
\label{subsec:problem_notation}

We assume a synchronized multi-sensor setup, consisting of a single LiDAR scan and a set of calibrated cameras $\mathcal{V}$ providing $360^\circ$ coverage (\eg{} surround-view systems). Let $\mathcal{P}=\{(\mathbf{x}_i,\mathbf{a}_i)\}_{i=1}^{N}$ denote the LiDAR point cloud, where $\mathbf{x}_i\in\mathbb{R}^3$ are 3D coordinates expressed in the LiDAR frame $\ell$ and $\mathbf{a}_i\in\mathbb{R}^{L}$ are raw point attributes (\eg{} intensity). The corresponding RGB images are $\mathcal{I}=\{\mathbf{I}^{v}\}_{v\in\mathcal{V}}$, where $\mathbf{I}^{v}\in\mathbb{R}^{H\times W\times 3}$ is the image captured by camera $v$. Each camera $v\in\mathcal{V}$ has intrinsics $\mathbf{Int}^{v}$ and an extrinsic transformation $\mathbf{Ext}^{v}_{\ell}$ mapping points from the LiDAR frame $\ell$ to the camera $v$. Using these calibration parameters, each LiDAR point $\mathbf{x}_i$ is projected into view $v$ as
\begin{equation}
\mathbf{u}^{v}_i = \pi^{v}(\mathbf{x}_i) \in \mathbb{R}^2,
\label{eq:projection}
\end{equation}
where $\pi^{v}(\cdot)$ denotes the perspective projection induced by $(\mathbf{Int}^{v},\mathbf{Ext}^{v}_{\ell})$. We define the visible subset $\mathcal{P}^v \subseteq \mathcal{P}$ as the points whose projections fall within the image bounds of $\mathbf{I}^v$ (and satisfy the positive depth constraint), establishing point--pixel correspondences used for VFM-to-LiDAR cross-modal distillation.

\subsection{Frustum Pooling: Point-to-Token Mapping}
\label{subsec:pooling}

The student ViT $\mathcal{G}_\phi(\cdot)$ operates on a regular grid of $N_p{=}(H/p){\times}(W/p)$ patch tokens with patch size $p$,
whereas the LiDAR backbone $\mathcal{L}_\theta(\cdot)$ produces unordered point features $\mathbf{f}_i\in\mathbb{R}^{C_{w}}$ for each LiDAR point and ${C_{w}}$ is LiDAR feature dimension. Using the projections $\mathbf{u}_i^v$, we define for each patch $t$ in view $v$ with ${v\in\mathcal{V}}$ a frustum set $\mathcal{N}^v(t)$ containing all visible points whose projections fall within the corresponding patch. A binary mask $\mathbf{M}^v \in {\{0,1\}}^{N_p}$ indicates whether a patch contains projected LiDAR points, \ie{} $\mathbf{M}_t^v=1$ if $\mathcal{N}^v(t)\neq\emptyset$ and $\mathbf{M}_t^v=0$ otherwise.
At the initial stage ($j{=}0$), where $j \in \{0,\ldots,{J-1}\}$, we replace the standard ViT patch embedding with tokens derived from the LiDAR backbone $\mathcal{L}_\theta$. For each patch $t$ in view $v$, point features within its frustum $\mathcal{N}^v(t)$ are linearly projected to the ViT embedding space and mean-pooled:
\begin{equation}
\mathbf{z}_t^v = \frac{1}{|\mathcal{N}^v(t)|} \sum_{i \in \mathcal{N}^v(t)} \mathbf{W}_p \mathbf{f}_i,
\label{eq:frustum_pooling}
\end{equation}
where $\mathbf{W}_p \in \mathbb{R}^{C_{\mathcal{S}} \times {C_w}}$. Patches without visible points ($\mathbf{M}_t^v = 0$) receive a learnable empty token $\mathbf{e} \in \mathbb{R}^{C_{\mathcal{S}}}$. The resulting token grid $\mathbf{Z}^v \in \mathbb{R}^{N_p \times C_{\mathcal{S}}}$, augmented with positional embeddings and a \texttt{[CLS]} token, is fed to the student ViT blocks. This initialization allows the transformer to operate directly in its native token space while preserving geometric structure via image-to-LiDAR projections. However, mean pooling treats all points equally and cannot distinguish foreground structure from background clutter within a frustum. We therefore introduce a content-aware aggregation mechanism at deeper stages.\looseness=-1

\subsection{Frustum Attention: Content-Aware 3D$\to$2D Injection}
\label{subsec:frustum_attention}

\begin{figure}[t!]
    \centering
    \includegraphics[width=\linewidth]{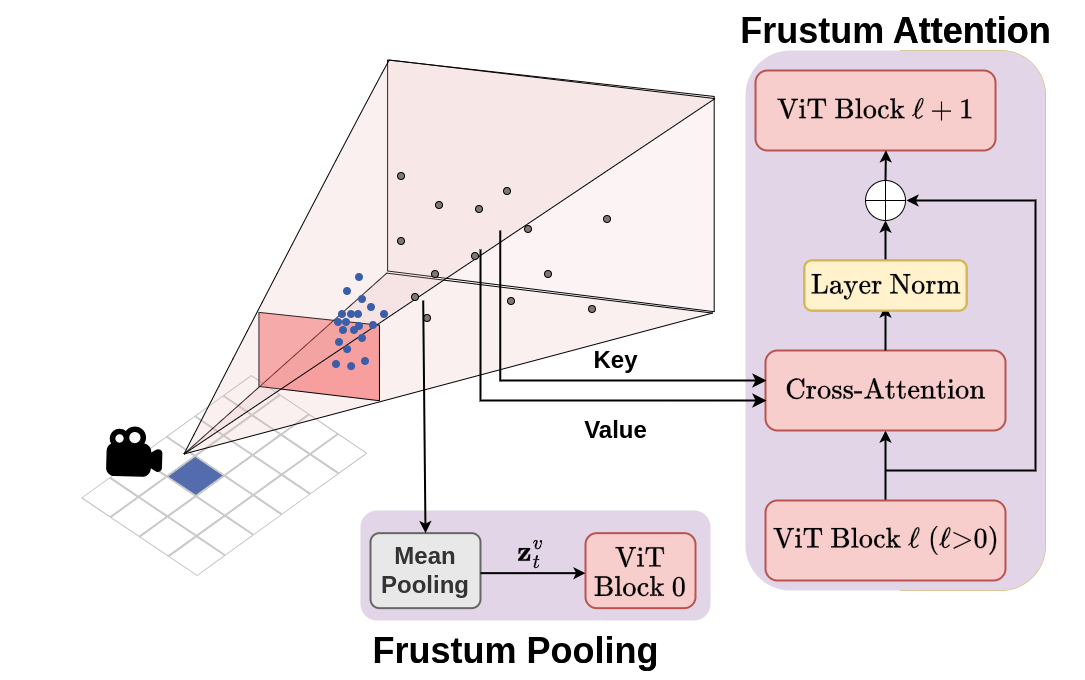} 
    \caption{\textbf{\emph{Frustum Pooling} and \emph{Frustum Attention}.} At the ViT input ($j{=}0$), point features are projected and mean-pooled into patch tokens; at deeper ViT blocks ($j{>}0$), each token attends to its frustum points via cross-attention.}
    \label{fig:frustum_both}
    \vspace{-3mm}
\end{figure}

At deeper blocks ($j{>}0$), patch tokens encode progressively contextual representations. We therefore introduce \emph{Frustum Attention (FA)}, a cross-attention mechanism that allows each token to selectively aggregate features from the set of 3D points within its corresponding frustum.
Given a patch token $\mathbf{z}_t$ and the set of point features $\{\mathbf{f}_i\}_{i \in \mathcal{N}^v(t)}$  (Fig.~\ref{fig:frustum_both}), we first map point features to the ViT embedding dimension using a stage-specific linear projection $\mathbf{W}_p^{(j)}$, yielding $\mathbf{g}_i=\mathbf{W}_p^{(j)}\mathbf{f}_i$ (we omit the superscript $(j)$ for clarity). We then compute Query ($\mathbf{Q}$), Key ($\mathbf{K}$), and Value ($\mathbf{V}$) projection matrices as:
\begin{equation}
\mathbf{Q}_t = \mathbf{W}_Q \mathbf{z}_t ;
\mathbf{K}_i = \mathbf{W}_K \mathbf{g}_i ;
\mathbf{V}_i = \mathbf{W}_V \mathbf{g}_i,
\label{eq:qkv}
\end{equation}
where $\mathbf{g}_i$=$\mathbf{W}_p\mathbf{f}_i$ maps point features to the ViT embedding space.
We learn separate projections at each stage; for readability we omit the stage index on $\mathbf{W}_p$, $\mathbf{W}_Q$, $\mathbf{W}_K$, $\mathbf{W}_V$.
Attention weights and aggregated context are computed as:
\begin{equation}
\begin{split}
\Delta\mathbf{z}_t = \sum_{i \in \mathcal{N}^v(t)} \mathrm{softmax}_{i \in \mathcal{N}^v(t)}\!\left(\frac{\mathbf{Q}_t^\top \mathbf{K}_i}{\sqrt{d}}\right)\,\mathbf{V}_i,
\label{eq:frustum_attention}
\end{split}
\end{equation}
where $d=C_{\mathcal{S}}/n_h$ is the per-head dimension and $n_h$ is the number of attention heads. The update is applied via residual addition with layer normalization $\mathrm{LN}(\cdot)$ to stabilise the injected cross-attention output: 
\begin{equation}
\mathbf{z}_t \leftarrow \mathbf{z}_t + \mathrm{LN}\!\left(\Delta\mathbf{z}_t\right).
\label{eq:gated_inject}
\end{equation}

The softmax in Eq. \ref{eq:frustum_attention} is computed independently within each frustum. Unlike static pooling in ViT, \emph{Frustum Attention (FA)} enables geometry injection conditioned on the token’s current semantic state, allowing selective aggregation of object-relevant structure while suppressing background noise.\looseness=-1

\subsection{Multi-Stage LiDAR Integration with Student ViT}
\label{subsec:fusion}
\emph{Frustum Pooling (FP)} and \emph{Frustum Attention (FA)} integrate the LiDAR backbone $\mathcal{L}_\theta$ at $J$ stages within the ViT blocks of the student $\mathcal{G}_\phi$, enabling multi-stage point features to be progressively mapped into patch-token representations. Let $D_{w}$ denote the total number of layers of $\mathcal{L}_\theta$; between successive stages, $\mathcal{L}_\theta$ advances by $D_{w}/J$ backbone layers alongside the corresponding ViT blocks, and the resulting point features are mapped to view-specific patch-token grids using the projections ${\mathbf{u}_i^v}$.

Unlike conventional VFM-to-LiDAR cross-modal distillation where a 3D backbone is required to approximate the teacher’s representation independently, $\mathcal{L}_\theta$ is used to supply geometry-conditioned features that complement the student ViT’s pretrained processing at multiple stages. The ViT retains responsibility for global reasoning and semantic abstraction through self-attention. Distillation gradients propagate through the Frustum modules and the ViT blocks into $\mathcal{L}_\theta$, guiding the LiDAR backbone to learn point-wise features that are compatible with the teacher-supervised token representations, rather than relying on direct regression from a standalone 3D encoder. \looseness=-1

\subsection{Visibility-Masked Distillation}
\label{subsec:masked_distillation}

For a given camera view $v$, the frozen teacher $\mathcal{T}$ processes the RGB image $\mathbf{I}^v$ to produce token representations $\mathbf{Z}^{\mathcal{T},v} = \{\mathbf{z}_\mathrm{cls}^{\mathcal{T},v},\, \mathbf{z}_t^{\mathcal{T},v}\}_{t=1}^{N_p}$, whereas the student ViT $\mathcal{G}_\phi$ is conditioned on patch tokens constructed from projected LiDAR features; it outputs $\mathbf{Z}^{\mathcal{S},v} = \{\mathbf{z}_\mathrm{cls}^{\mathcal{S},v},\, \mathbf{z}_t^{\mathcal{S},v}\}_{t=1}^{N_p}$.
A learnable projection $\psi(\cdot): \mathbb{R}^{C_{\mathcal{S}}} \rightarrow \mathbb{R}^{C_\mathcal{T}}$ maps student features to the teacher embedding space, where $C_\mathcal{T} $ is the embedding dimension of the frozen teacher $\mathcal{T}$.

\noindent\textbf{Distillation Objective:} Feature alignment is enforced via
\begin{equation}
\mathcal{L}_\mathrm{dist}(\mathbf{p}, \mathbf{q})
= \tfrac{1}{2}\bigl(\mathcal{L}_\mathrm{cos}(\mathbf{p}, \mathbf{q}) + \mathcal{L}_\mathrm{sl1}(\mathbf{p}, \mathbf{q})\bigr),
\label{eq:dist_loss}
\end{equation}
where $\mathbf{p}$ and $\mathbf{q}$ denote any two embeddings being aligned, $\mathcal{L}_\mathrm{cos}=1-\cos(\cdot)$ and $\mathcal{L}_\mathrm{sl1}$ denotes smooth-$\ell_1$. This combination constrains both directional consistency and feature magnitude.

\noindent\textbf{Global Alignment:} We align the \texttt{[CLS]} tokens through
\begin{equation}
\mathcal{L}_\mathrm{cls} = \mathcal{L}_\mathrm{dist}\left(\psi(\mathbf{z}_\mathrm{cls}^{\mathcal{S},v}), \mathbf{z}_\mathrm{cls}^{\mathcal{T},v}\right).
\label{eq:cls_loss}
\end{equation}

\noindent\textbf{Visibility-Masked Local Alignment:} Patch-level supervision is applied only to tokens supported by projected LiDAR geometry using the occupancy mask $\mathbf{M}^v$:

\begin{equation}
\mathcal{L}_\mathrm{patch}
= \frac{1}{\sum_{t}\mathbf{M}_t^v}\sum_{t=1}^{N_p} \mathbf{M}_t^v\,\mathcal{L}_\mathrm{dist}\!\left(\psi(\mathbf{z}_t^{\mathcal{S},v}),\;\mathbf{z}_t^{\mathcal{T},v}\right),
\label{eq:patch_loss}
\end{equation}

\noindent thereby avoiding penalization of regions without geometric evidence. The final objective combines global and local terms:
\begin{equation}
\mathcal{L} = \tfrac{1}{2}(\mathcal{L}_\mathrm{cls} + \mathcal{L}_\mathrm{patch}).
\label{eq:total_loss}
\end{equation}

This visibility-masked distillation ensures that supervision remains physically grounded in observed 3D structure, preventing spurious alignment in image-only regions.\looseness=-1

\subsection{Finetuning via Masked Bilinear Sampling}
\label{subsec:finetune}

During finetuning, $\mathcal{L}_\theta$ features are integrated into $\mathcal{G}_\phi$ at all $J$ stages and across all $\mathcal{V}$ views. At the final stage, ViT tokens are lifted to the point domain and concatenated with point-wise features from $\mathcal{L}_\theta$, yielding geometry-aware and semantically enriched point representations.

\begin{figure}[!t]
\centering
\includegraphics[width=0.8\linewidth, trim=30pt 60pt 80pt 40pt, clip]{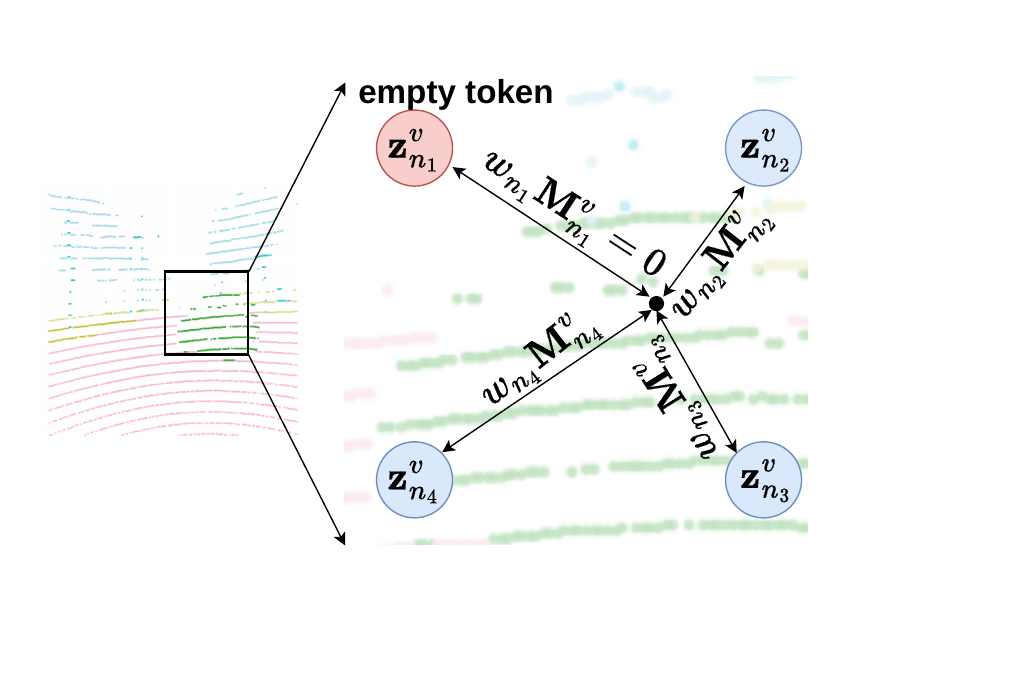} 
\caption{\textbf{Masked Bilinear Sampling for 2D$\rightarrow$3D Lifting.} Empty patches (red) are suppressed by the occupancy mask $\mathbf{M}^v_n$ and the contributions from valid tokens (blue) are renormalised, preventing spurious features in lifted 3D points.}
\label{fig:masked_bilinear}
\end{figure}

\noindent\textbf{Masked Bilinear Sampling:} Naive bilinear interpolation at a point’s projection $\mathbf{u}_i^v$ aggregates all four neighbouring tokens, including empty patches, introducing spurious semantics. We instead mask interpolation using the occupancy map $\mathbf{M}^v$ (Fig.~\ref{fig:masked_bilinear}):

\begin{equation}
\mathbf{r}_i^v
= \frac{\sum_{n \in \Omega(\mathbf{u}_i^v)} w_n \cdot \mathbf{z}_n^v \cdot \mathbf{M}_n^v}{\sum_{n \in \Omega(\mathbf{u}_i^v)} w_n \cdot \mathbf{M}_n^v + \epsilon},
\label{eq:masked_bilinear}
\end{equation}

\noindent where $\Omega(\mathbf{u}_i^v)$ denotes the four grid neighbors and $w_n$ are bilinear weights. Empty tokens are suppressed and weights renormalised over valid support only.

\noindent\textbf{Cross-view Aggregation and Segmentation:} 
A single LiDAR point $\mathbf{x}_i$ may be visible in multiple cameras. For each such view $v$ where $i \in \mathcal{P}^v$, masked bilinear sampling produces a lifted feature $\mathbf{r}_i^v$. These are averaged across all views that observe the point $\mathbf{r}_i$=$\mathrm{Mean}(\{\mathbf{r}_i^v \mid v:i \in \mathcal{P}^v\})$. 
The aggregated view feature $\mathbf{r}_i$ is then passed to a segmentation head for per-point prediction $\hat{y}_i = \mathrm{SegHead}([\mathbf{f}_i \;\|\; \mathbf{r}_i])$,.

\noindent Overall, \coolname{} preserves the teacher’s architectural prior by retaining a same-family ViT during distillation, while \emph{Frustum Pooling/Attention} and masked lifting enforce geometrically grounded cross-modal alignment.\looseness=-1

%% file: Section/experiments.tex
\section{Experimental Setups}
\label{sec:experiments}

\subsection{Datasets}

We pretrain \coolname{} on five large-scale autonomous driving datasets with 78,791 LiDAR scans and 341,645 camera-LiDAR pairs, spanning diverse sensor configurations, viewpoints, and driving environments.

\noindent\textbf{\emph{nuScenes}} \cite{caesar2020nuscenes}: 28,130 keyframes with 16 semantic classes, captured using a 32-beam LiDAR ($360^\circ$) and 6 surround-view cameras in Boston and Singapore, yielding 168,780 camera-LiDAR pairs.

\noindent\textbf{\emph{SemanticKITTI}}~\cite{behley2019semantickitti}: 19,130 frames with dense point-wise annotations over 19 classes, acquired from a 64-beam LiDAR (KITTI odometry sequences), providing 19,130 camera-LiDAR pairs.

\noindent\textbf{\emph{PandaSet}}~\cite{10.1109/ITSC48978.2021.9565009}: Two splits are used. PandaSet-64 (3,920 frames) employs a 64-beam LiDAR with 6 cameras (23,520 pairs); PandaSet-GT (3,920 frames) uses roughly 150 channels of forward-facing LiDAR with 3 front cameras (11,760 pairs). Both provide 17 semantic classes.

\noindent\textbf{\emph{Waymo Open}}~\cite{sun2020scalability}: Following MuDDoS~\cite{Michele_2025_BMVC}, we use a 23,691-frame subset annotated over 22 classes. Data are captured with a 64-beam LiDAR and 5 cameras across diverse U.S. urban/suburban environments, contributing 118,455 camera-LiDAR pairs.

Overall, the pretraining corpus spans a broad range of LiDAR configurations, including forward-facing, high-angular-resolution sensing (\ie{} Hesai PandarGT, roughly 150 channels, 60-degree HFOV, 10 Hz, up to 300 m) and mechanical spinning 360-degree LiDARs with 32–64 beams (\ie{} \emph{nuScenes} at 32 beams; \emph{SemanticKITTI}, \emph{Waymo Open}, and \emph{PandaSet-64} at 64 beams), with capture rates up to 20 Hz and ranges from about 100 to 300 m, alongside heterogeneous multi-camera rigs, providing substantial geometric and cross-sensor diversity for self-supervised LiDAR learning.

\subsection{Evaluation Protocols}

We evaluate \coolname{} under four complementary protocols assessing representation quality, label efficiency, and cross-domain generalization. All results are reported in mIoU for consistent comparison across datasets and tasks.

\noindent\textbf{Linear probing (LP):} Following \cite{puy2024three}, the pretrained backbone is frozen and a lightweight head (BatchNorm + linear classifier) is trained on data. We report mIoU on the official validation splits of \emph{nuScenes}, \emph{SemanticKITTI}, \emph{PandaSet-64}, and \emph{PandaSet-GT}.

\noindent\textbf{Label-Efficient Finetuning:} The full model is finetuned using  few-shot finetuning protocols on 1\% and 10\% labels on \emph{nuScenes} and 1\% on \emph{SemanticKITTI}, measuring low-data regimes performance~\cite{puy2024three}.

\noindent\textbf{Full Finetuning:} All parameters are trained on 100\% \emph{nuScenes} labels.

\noindent\textbf{Domain Adaptation (DA):} We follow unsupervised domain adaptation protocol \cite{Michele_2025_BMVC}. The pretrained backbone is frozen and an MLP head is trained on a labeled source domain and evaluated on an unlabeled target domain over 10 shared classes. We report N$\rightarrow$K, K$\rightarrow$N, N$\rightarrow$W, and W$\rightarrow$N (N: \emph{nuScenes}, K: \emph{SemanticKITTI}, W: \emph{Waymo Open}).

\begin{table}[!t]
\centering
\caption{\textbf{Comparison with prior camera-LiDAR distillation methods.} mIoU (\%) on \emph{nuScenes} (LP, 1\%, 10\%, 100\%) and \emph{SemanticKITTI} (1\%). }
\label{tab:main_comparison}
\small
\begin{tabular}{P{1.9cm} P{0.9cm} P{0.4cm} P{0.4cm} P{0.4cm} P{0.4cm} C{1.1cm}}
\toprule
\multirow{2}{*}{Method} & \multirow{2}{*}{Pretrain} & \multicolumn{4}{c}{nuScenes} & S.KITTI \\
\cmidrule(lr){3-6} \cmidrule(lr){7-7}
 & & LP & 1\% & 10\% & 100\% & 1\% \\
\midrule
PPKT~\cite{liu2021learning} & nuSc. & \textit{35.9} & \textit{37.8} & \textit{60.3} & \textit{74.5} & \textit{44.0} \\
SLidR~\cite{sautier2022image} & nuSc. & \textit{38.8} & \textit{38.3} & \textit{59.8} & \textit{74.8} & \textit{44.6} \\
ST-SLidR~\cite{mahmoud2023self} & nuSc. & \textit{40.5} & \textit{40.8} & \textit{60.8} & \textit{75.1} & \textit{44.7} \\
Seal~\cite{liu2023segment} & nuSc. & \textit{45.0} & \textit{45.8} & \textit{63.0} & \textit{75.6} & \textit{46.6} \\
\midrule
LargeAD~\cite{kong2025largead} & Multi & \textit{48.7} & \textit{49.2} & \textit{64.8} & \textit{75.9} & \textit{51.7} \\
ScaLR~\cite{puy2024three} & Multi & 67.8 & 50.7 & \textbf{69.2} & \textbf{78.4} & {56.8} \\
\rowcolor{roworange} \textbf{\coolname{}} & Multi & \textbf{71.7} & \textbf{53.5} & \textbf{69.2} & 77.3  & \textbf{57.0} \\
\bottomrule
\end{tabular}
\vspace{-1.0em}
\end{table}

\begin{table}[!t]
\centering
\caption{\textbf{Unsupervised domain adaptation.} mIoU (\%) on 10 shared classes for four cross-sensor pairs. Mod.: modality (U: uni, M: multi). ST: self training with target dataset. $^*$: pretrained on five datasets (incl. \emph{Waymo Open}). \coolname{} (4-dataset pretraining, no \emph{Waymo Open}) surpasses MuDDoS trained on five datasets, indicating strong generalization to unseen sensors.}
\label{tab:domain_adaptation}
\setlength{\tabcolsep}{3pt}
\small
\begin{tabular}{P{2.85cm} C{0.55cm} C{0.8cm} C{0.8cm} C{0.7cm} C{0.8cm} C{0.7cm}}
\toprule
Method & Mod. & N$\rightarrow$K & K$\rightarrow$N & N$\rightarrow$W & W$\rightarrow$N & Avg. \\
\midrule
\rowcolor{rowgray} Target oracle & - & 72.4 & 83.8 & 83.4 & 83.8 & 80.9 \\
Source only & - & 44.6 & 55.1 & 37.1 & 64.6 & 50.4 \\
\midrule
PL~\cite{morerio2018minimalentropy} & U & 30.0 & 29.0 & 31.9 & 22.3 & 28.3 \\
CosMix~\cite{saltori2022cosmix} & U & 30.6 & 29.7 & 31.5 & 30.0 & 30.5 \\
MM2D3D~\cite{cardace2023exploiting} & M & 32.9 & 33.7 & 34.1 & 37.5 & 34.6 \\
Adapt-SAM~\cite{peng2024learning} & M & 48.5 & 42.9 & 44.9 & 48.2 & 46.1 \\
MuDDoS wo ST~\cite{Michele_2025_BMVC} & M & 41.8 & 53.3 & 54.1 & 53.0 & 50.6 \\
MuDDoS w ST~\cite{Michele_2025_BMVC} & M & {52.1} & {66.4} & {69.1} & {70.5} & {64.5} \\
\textbf{\coolname{} wo ST} & M & {53.7} & {66.7} & {58.4} & {66.8} & {61.4} \\
\rowcolor{roworange} \textbf{\coolname{}$^*$ wo ST} & M & \textbf{57.2} & \textbf{68.6} & \textbf{70.6} & \textbf{70.5} & \textbf{66.7} \\
\bottomrule
\end{tabular}
\end{table}
\subsection{Fixed-Camera Systems}
For downstream evaluation, we replace per-frame calibrations with a fixed dataset-level camera rig, selecting the most frequent intrinsic configuration. For \emph{nuScenes}, \emph{SemanticKITTI}, and \emph{PandaSet}, we adopt a standardized 6-camera rig based on the dominant \emph{nuScenes} configuration. For \emph{Waymo Open}, we augment its 5-camera setup with three rear virtual cameras (front extrinsics rotated 180$^\circ$ about the $z$-axis) to achieve full 360$^\circ$ coverage. This allows a single pretrained model to transfer across diverse datasets.

\subsection{Implementation Details}

The 3D backbone $\mathcal{L}_\theta$ is a 48-layer WaffleIron~\cite{Puy_2023_ICCV} with 768-dimensional features. The student $\mathcal{G}_\phi$ is a DINOv2 ViT-B/14~\cite{oquab2024dinov} (768-dim) initialized from pretrained weights, and the frozen teacher $\mathcal{T}$ is DINOv2 ViT-L/14 (1024-dim); a linear projector maps student features to the teacher dimension. \emph{Frustum Attention} is inserted every 4th ViT layer with 12 attention heads.

We optimize all trainable parameters with AdamW using an initial learning rate of $5\times10^{-4}$, cosine-decayed to $10^{-6}$ with 10\% warmup and weight decay 0.03. Training runs for 25 epochs on 8 NVIDIA H100 GPUs with a total batch size of 16 and gradient clipping at $1.0$. During distillation, a single camera view is randomly sampled per scene; the teacher processes the corresponding RGB image, while the student is conditioned on the LiDAR-derived tokens from the same view. Images are tightly cropped to the 2D support of projected LiDAR points and resized to $518\times518$, yielding a $37\times37$ token grid for ViT-14. During finetuning, all available views are used.

Point cloud augmentations include random rotation about the $z$-axis, random XY flips, and isotropic scale jitter ($\pm 0.1$). The distillation loss combines cosine similarity and smooth-$\ell_1$ over the \texttt{[CLS]} token and visibility-masked patch tokens. Following~\cite{sariyildiz2025dune}, teacher tokens are standardised to zero mean and unit variance before loss computation, with separate EMA running statistics for \texttt{[CLS]} and patch tokens synchronised across GPUs, and cosine-scheduled momentum from $\beta_0=1.0$ to $\beta_{\mathrm{end}}=0.001$.

%% file: Section/results.tex
\section{Results}
\label{sec:results}

\begin{table}[!t]
\centering
\caption{\textbf{Scaling pretraining data.} Linear probing mIoU (\%) under varying pretraining data: 169k  (\emph{nuScenes}), 223k  (\emph{nuScenes} + \emph{SemanticKITTI} + \emph{PandaSet-64} + \emph{PandaSet-GT}), and 342k (+ \emph{Waymo Open}). \coolname{} consistently outperforms ScaLR, with gains increasing as more data is used (N: \emph{nuScenes}, S: \emph{SemanticKITTI}, P: \emph{PandaSet}).}
\label{tab:cross_dataset}
\begin{tabular}{P{1.10cm} C{1.0cm} C{0.7cm} C{0.7cm} C{0.7cm} C{0.7cm}}
\toprule
& & \multicolumn{4}{c}{Downstream \& Test Dataset} \\
\cmidrule(lr){3-6}
Method & \#Data & N & S & P-64 & P-GT \\
\midrule
\multicolumn{6}{l}{\cellcolor{lightgray}\textit{Pretraining with DINO-ViT-S/8 and linear probing}} \\
ScaLR~\cite{puy2024three}  & 169k  & 54.4 & 28.8 & 26.9 & 25.2 \\
\coolname{} & 169k  & \textbf{56.4} & \textbf{40.9} & \textbf{27.6} & \textbf{29.3} \\
\arrayrulecolor{rulegray}\midrule\arrayrulecolor{black}
ScaLR~\cite{puy2024three}  & 223k  & 54.6 & \textbf{50.6} & \textbf{33.1} & 32.3 \\
\coolname{} & 223k  & \textbf{56.8} & 48.2 & 31.4 & \textbf{33.4} \\
\midrule
\multicolumn{6}{l}{\cellcolor{lightgray}\textit{Pretraining with DINOv2-ViT-L/14 and linear probing}} \\
ScaLR~\cite{puy2024three}  & 169k  & 67.8 & 43.1 & \textbf{33.9} & 29.9 \\
\coolname{} & 169k  & \textbf{68.9} & \textbf{48.8} & 32.3 & \textbf{32.8} \\
\arrayrulecolor{rulegray}\midrule\arrayrulecolor{black}
ScaLR~\cite{puy2024three}  & 223k  & 67.8 & 55.8 & 37.9 & 34.5 \\
\coolname{} & 223k  & \textbf{71.7} & \textbf{57.0} & \textbf{39.9} & \textbf{40.6} \\
\arrayrulecolor{rulegray}\midrule\arrayrulecolor{black}
ScaLR~\cite{puy2024three}  & 342k & 67.5 & 56.8 & 38.7 & 35.9 \\
\coolname{} & 342k & \textbf{70.9} & \textbf{60.1} & \textbf{42.0} & \textbf{42.6} \\
\bottomrule
\end{tabular}
\end{table}

\begin{figure}[t]
    \centering
    \includegraphics[width=0.8\columnwidth]{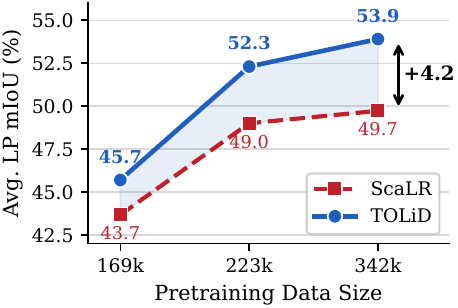}
    \caption{Effect of scaling pretraining data. Average mIoU across \emph{nuScenes},
    \emph{SemanticKITTI}, \emph{PandaSet-64}, and \emph{PandaSet-GT}. \coolname{} exhibits
    stronger scaling with data diversity, widening the gap over
    ScaLR~\cite{puy2024three} from +2.0\% at 169k  frames to +4.2\% at 342k
    frames.}
    \label{fig:scaling}
    \vspace{-1.0em}
\end{figure}

\begin{table*}[!t]
    \centering
\caption{\textbf{Robustness probing} on the nuScenes-C split of the Robo3D benchmark \cite{kong2023robo3d}, which demonstrates eight real-world corruption types from severe weather, external disturbances, and internal sensor failures. Each corruption is evaluated across different severity levels, and mIoU (\%) is reported per corruption type and averaged.
$^*$ denotes the multi-dataset pretrained model evaluated under both \textbf{LP} and \textbf{Full} finetuning protocols. The substantial LP improvement reflects the quality of \coolname{} frozen representations, which yield deployment-ready features that remain robust under sensor corruptions.}
    \vspace{-0.1cm}
\label{tab:robustness}
\resizebox{0.8\linewidth}{!}{
    \begin{tabular}{c|r|p{0.8cm}<{\centering}p{0.8cm}<{\centering}p{0.8cm}<{\centering}p{0.8cm}<{\centering}p{0.8cm}<{\centering}p{0.8cm}<{\centering}p{0.8cm}<{\centering}p{0.8cm}<{\centering}|p{0.8cm}<{\centering}}
    \hline 
    \textbf{\#} & \textbf{Method} & \textbf{Fog} & \textbf{Wet} & \textbf{Snow} & \textbf{Move} & \textbf{Beam} & \textbf{Cross} & \textbf{Echo} & \textbf{Sensor} & \textbf{Avg} \\
    \cline{1-11}
        \multirow{7}{*}{\rotatebox[origin=c]{90}{\textbf{LP}}}
    & PPKT~\cite{liu2021learning}  & 30.65 & 35.42 & 28.12 & 29.21 & 32.82 & 19.52 & 28.01 & 20.71 & 28.06 \\
    & SLidR~\cite{sautier2022image}  & 34.88 & 38.09 & 32.64 & 26.44 & 33.73 & 20.81 & 31.54 & 21.44 & 29.95 \\
    & Seal~\cite{liu2023segment}  & 37.33 & 42.77 & 29.93 & 37.73 & 40.32 & 20.31 & 37.73 & 24.94 & 33.88 \\
    & Superflow~\cite{xu20244d}  & 37.59 & 43.42 & 37.60 & 39.57 & 41.40 & 23.64 & 38.03 & 26.69 & 35.99 \\
    & LargeAD~\cite{kong2025largead}  & 40.54 & 43.26 & 37.92 & 38.27 & 40.27 & 25.67 & 39.26 & 30.62 & 36.98 \\
    & ScaLR~\cite{puy2024three}  & 59.87 & 66.47 & 57.77 & 46.13 & 58.20 & 45.97 & 53.13 & 40.53 & 53.51 \\
    \cline{2-11} \noalign{\vskip 2pt}
    & ScaLR$^{*}$~\cite{puy2024three}  & 59.82 & 66.39 & 57.75 & 46.14 & 58.08 & 46.01 & 53.07 & 40.42 & 53.46 \\
    \rowcolor{roworange} & {\textbf{\coolname{}}$^{*}$}  & \textbf{67.28} & \textbf{71.41} & \textbf{66.28} & \textbf{54.14} & \textbf{63.73} & \textbf{63.52} & \textbf{61.05} & \textbf{43.28} & \textbf{61.34} \\
    \hline
    \hline
    \multirow{9}{*}{\rotatebox[origin=c]{90}{\textbf{Full}}} 
    & PPKT~\cite{liu2021learning}  & 64.01 & 72.18 & 59.08 & 57.17 & 63.88 & 36.34 & 60.59 & 39.57 & 56.60 \\
    & SLidR~\cite{sautier2022image}  & 65.41 & 72.31 & 56.01 & 56.07 & 62.87 & 41.94 & 61.16 & 38.90 & 56.83 \\
    & Seal~\cite{liu2023segment}  & \textbf{72.66} & 74.31 & 66.22 & \textbf{66.14} & 65.96 & 57.44 & 59.87 & 39.85 & 62.81 \\
    & Superflow~\cite{xu20244d}  & 70.32 & 75.77 & 65.41 & 61.05 & 68.09 & 60.02 & 58.36 & 50.41 & 63.68 \\
    & LargeAD~\cite{kong2025largead}  & 71.95 & 72.47 & 67.28 & {65.29} & 67.49 & 59.42 & 61.38 & 42.46 & 63.47 \\
    & ScaLR~\cite{puy2024three}  & 70.8 & {77.2} & 67.1 & 55.9 & {70.0} & {65.7} & 63.9 & {51.1} & 65.20 \\
    \cline{2-11} \noalign{\vskip 2pt}
    & ScaLR$^{*}$~\cite{puy2024three}  & 72.2 & \textbf{77.9} & 69.1 & 57.4 & \textbf{70.1} & {62.7} & 64.0 & \textbf{52.2} & 65.70 \\
    \rowcolor{roworange} & {\textbf{\coolname{}}$^{*}$} &  70.42 & {77.21} & \textbf{71.70} &57.89 & 69.37 & \textbf{67.07} & \textbf{65.83} & 46.27 & \textbf{65.72} \\
    \hline

    \end{tabular}
    
}
\vspace{-1.0em}
\end{table*}

\begin{table}[!t]
\centering
\caption{\textbf{Ablation study evaluating the effect of sampling strategy and frustum aggregation} on \emph{nuScenes} and \emph{PandaSet-64}. Top: effect of masked vs naive bilinear sampling. Bottom: \emph{Frustum Attention vs Pooling} across fusion stages. LP mIoU (\%)}
\label{tab:ablation_combined}
\small
\begin{tabular}{P{2.5cm} P{2.1cm} C{1cm} C{1.2cm}}
\toprule
 Ablation Type & Setting & \emph{nuScenes} & \emph{Pand.\ 64} \\
\midrule
\multirow{2}{*}{Sampling Strategy} & Naive bilinear & 71.3 & 39.7 \\
& Masked bilinear & \textbf{71.7} & \textbf{39.9} \\
\midrule
\multirow{2}{*}{Frustum Aggregation} & Frustum Pool & 56.0 & 27.4 \\
& Frustum Attn. & \textbf{56.4} & \textbf{27.6} \\
\bottomrule
\end{tabular}
\vspace{-1.0em}
\end{table}

\noindent\textbf{Linear Probing and Label-Efficient Finetuning:} Tab. \ref{tab:main_comparison} reports results on \emph{nuScenes} and \emph{SemanticKITTI}. Baseline results are taken directly from the corresponding publications. 
\coolname{} is pretrained on four heterogeneous datasets: \emph{nuScenes}, \emph{SemanticKITTI}, \emph{PandaSet-64}, and \emph{PandaSet-GT}.
Under linear probing, \coolname{} attains 71.7\% mIoU on \emph{nuScenes}, surpassing the previous best (ScaLR \cite{puy2024three}, 67.8\%) by +3.9\%. Since linear probing evaluates frozen representations, this margin reflects a substantive improvement in feature quality. Retaining the VFM within the distillation loop yields 3D embeddings that are more semantically structured and linearly separable than those obtained through cross-architectural distillation.

\noindent\textbf{Cross-Sensor Domain Adaptation:} Tab. \ref{tab:domain_adaptation} reports unsupervised domain adaptation under the MuDDoS protocol \cite{Michele_2025_BMVC}. 
Without any self-training (ST) on the target domain, a step that MuDDoS~\cite{Michele_2025_BMVC} relies on for its best results, \coolname{} pretrained on all five datasets achieves the best performance on all four cross-sensor pairs, averaging 66.7\% mIoU (+2.2\% over MuDDoS~\cite{Michele_2025_BMVC}).
The largest improvement occurs on the challenging N$\rightarrow$K setting, increasing from 52.1\% to {57.2\%} mIoU (+5.1\%), corresponding to the sensor gap (32-beam \emph{nuScenes} $\rightarrow$ 64-beam \emph{SemanticKITTI}). 
Notably, even without \emph{Waymo Open} pretraining, \coolname{} already  surpasses MuDDoS~\cite{Michele_2025_BMVC} without self-training on N$\rightarrow$W (58.4\% vs.\ 54.1\%) and W$\rightarrow$N (66.8\% vs.\ 53.0\%), demonstrating that \coolname{} generalizes to unseen sensors during pretraining phase.
These gains are obtained using our frozen-backbone and lightweight MLP head, confirming that self-supervised representations learned via distillation over compatible patch-token representations are intrinsically robust to cross-sensor domain shifts.

\input{Section/qualitative}
\noindent\textbf{Cross-Dataset Generalization.} Tab.~\ref{tab:cross_dataset} evaluates how pretraining data scale affects linear probing performance across four datasets with heterogeneous LiDAR sensors. With a DINOv2-ViT-L/14 teacher, \coolname{} outperforms ScaLR~\cite{puy2024three} on the majority of settings, and the average gap widens with increasing data volume. At 342k frames, \coolname{} achieves 70.9\%, 60.1\%, 42.0\%, and 42.6\% mIoU on \emph{nuScenes}, \emph{SemanticKITTI}, \emph{PandaSet-64}, and \emph{PandaSet-GT} respectively, outperforming ScaLR by +3.4\%, +3.3\%, +3.3\%, and +6.7\% mIoU. The gain on \emph{PandaSet-GT} is particularly notable, as this forward-facing sensor presents the largest domain gap from the predominantly 360$^\circ$ pretraining distributions. At 223k  frames, \coolname{} already reaches its peak \emph{nuScenes} score of 71.7\% mIoU (+3.9\% over ScaLR) and 40.6\% on \emph{PandaSet-GT} (+6.1\%), while scaling to 342k frames further boosts \emph{SemanticKITTI} by +3.1\% and both \emph{PandaSet} splits by at least +2.0\% mIoU, at a modest cost on \emph{nuScenes}. Fig.~\ref{fig:scaling} confirms this trend: the average LP gap widens from +2.0\% mIoU at 169k  frames to +4.2\% at 342k frames, indicating that the coupled LiDAR-ViT architecture leverages geometric diversity from heterogeneous sensors more effectively than feature-level alignment, which saturates earlier.  Fig.~\ref{fig:qualitative} shows qualitative LP comparisons on \emph{SemanticKITTI}, where \coolname{} produces more semantically consistent predictions than ScaLR~\cite{puy2024three}. This is encouraging for robotic fleets that continually accumulate unlabeled multi-sensor data, as \coolname{} can translate such data growth into improved representations without additional annotation effort.

\noindent\textbf{Robustness to Sensor Corruptions:} Tab.~\ref{tab:robustness} evaluates robustness on the Robo3D benchmark~\cite{kong2023robo3d}, which simulates eight corruption types spanning severe weather (fog, wet ground, snow), external disturbances (motion blur, beam missing), and internal sensor failure (crosstalk, incomplete echo, cross-sensor). 
Under LP, \coolname{} achieves 61.34\% mIoU compared to 53.46\% for ScaLR~\cite{puy2024three}, improving across all eight corruptions. Gains are consistent across corruption categories, with the largest improvement on crosstalk (+17.51\%), where multi-path interference introduces spurious points. 
Under full finetuning both methods converge (65.72\% vs.\ 65.70\%), confirming that the advantage is specific to frozen representations. 
This distinction is practically relevant: \coolname{} enables faster adaptation to new robotic platforms by training only a lightweight head, while providing strong robustness under real-world sensor perturbations.

\noindent\textbf{Ablation Studies:}
 Tab.~\ref{tab:ablation_combined} presents an ablation study on two key design choices. First, we show Masked bilinear sampling (Eq.~\ref{eq:masked_bilinear}) restricts interpolation to geometrically supported tokens and re-normalizes over valid support only, improving LP mIoU from 71.3\% to 71.7\% on \emph{nuScenes} and from 39.7\% to 39.9\% on \emph{PandaSet-64}. Naive bilinear interpolation contaminates lifted per-point representations. 
 Second, we compare \emph{Frustum Attention} against \emph{Frustum Pooling} at the deeper fusion stages (Fig.~\ref{fig:frustum_both}) using a small-scale setting (DINO ViT-S/8, WI-256, \emph{nuScenes} only).
 Both variants share \emph{Frustum Pooling} at layer~0 for initialization and differ only at the three subsequent injection points. \emph{Frustum Attention} raises LP mIoU from 56.0\% to 56.4\% on \emph{nuScenes} and from 27.4\% to 27.6\% on \emph{PandaSet-64}, confirming that content-aware cross-attention extracts richer cues than mean pooling.

%% file: Section/qualitative.tex
\begin{figure}[!t]
    \centering
    \makebox[0.32\linewidth]{\small GT}%
    \hfill
    \makebox[0.32\linewidth]{\small ScaLR~\cite{puy2024three}}%
    \hfill
    \makebox[0.32\linewidth]{\small \coolname{}}%
    \vspace{2pt}

    \includegraphics[width=0.32\linewidth]{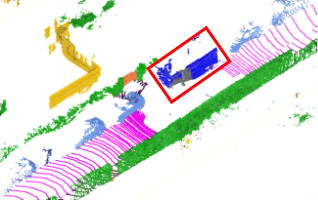}%
    \hfill
    \includegraphics[width=0.32\linewidth]{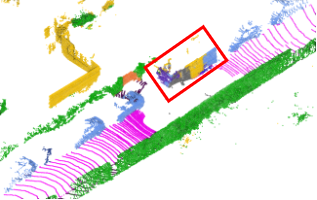}%
    \hfill
    \includegraphics[width=0.32\linewidth]{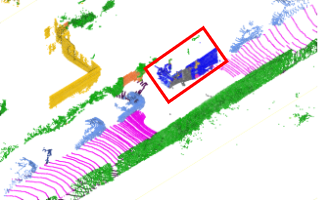}%

    \vspace{2pt}

    \includegraphics[width=0.32\linewidth]{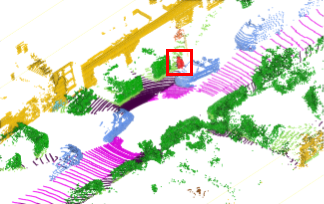}%
    \hfill
    \includegraphics[width=0.32\linewidth]{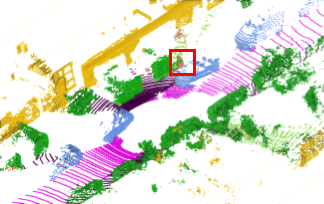}%
    \hfill
    \includegraphics[width=0.32\linewidth]{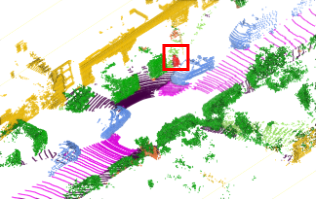}%
    \caption{Qualitative results on \emph{SemanticKITTI} (linear probing). \coolname{} produces more semantically consistent predictions than ScaLR~\cite{puy2024three}.}
    \label{fig:qualitative} 
\end{figure}

%% file: Section/conclusion.tex
\section{Conclusion}

We introduced \coolname{}, a self-supervised cross-modal distillation framework for LiDAR representation learning that addresses a key limitation of prior VFM-to-LiDAR methods: the need to align sparse 3D features directly to a frozen visual teacher across a substantial architectural mismatch. Instead of distilling into a standalone heterogeneous 3D encoder, \coolname{} couples a LiDAR backbone with a student ViT from the same architectural family as the frozen teacher, enabling supervision in a compatible patch-token space. Through \emph{Frustum Pooling} and \emph{Frustum Attention}, projected LiDAR features are converted into geometry-aware tokens, while visibility-masked distillation and masked bilinear sampling ensure that both supervision and feature lifting remain grounded in valid 3D support. Extensive experiments across five heterogeneous pretraining datasets show that \coolname{}  learns stronger and more scalable representations, consistently improving linear probing, label-efficient finetuning, cross-dataset generalization, cross-sensor domain adaptation, and robustness to sensor corruptions. In particular, its advantages are most pronounced when transferring frozen backbones with lightweight heads, demonstrating that closing the architecture gap during distillation leads to more transferable, semantically structured, and deployment-ready LiDAR features for real-world 3D perception.

\section*{Acknowledgment}
The authors would like to acknowledge support from the CSIRO’s Embodied AI cluster and partial support from the Australian Research Council (ARC) Discovery Project Grant DP250103634.